\pdfoutput=1

\documentclass[11pt]{article}

\usepackage[]{ACL2023}

\usepackage{times}
\usepackage{latexsym}
\usepackage{xcolor}
\usepackage[T1]{fontenc}

\usepackage[utf8]{inputenc}

\usepackage{microtype}

\usepackage{inconsolata}
\usepackage{amsfonts} 
\usepackage{adjustbox}
\usepackage{comment}
\usepackage{subcaption}
\usepackage{caption}
\usepackage{booktabs}
\usepackage{multirow}
\usepackage{textcomp}
\newcommand{\modelname}[0]{HypEmo}
\newcommand{\wsmodelname}[0]{HypEmo }

\renewcommand{\textuparrow}{$\uparrow$}
\renewcommand{\textdownarrow}{$\downarrow$}
\title{Label-Aware Hyperbolic Embeddings\\for Fine-grained Emotion Classification}

\author{Chih-Yao Chen\textsuperscript{1}, Tun-Min Hung\textsuperscript{2}, Yi-Li Hsu\textsuperscript{2}, Lun-Wei Ku\textsuperscript{2}\\
  \textsuperscript{1}UNC Chapel Hill, \textsuperscript{2}Institute of Information Science, Academia Sinica \\
\texttt{\textsuperscript{1}cychen@cs.unc.edu} \\
\texttt{\textsuperscript{2}{\string{allenhung,yili.hsu,lwku\string}}@iis.sinica.edu.tw} \\}

\begin{document}
\maketitle
\begin{abstract}
Fine-grained emotion classification (FEC) is a challenging task. Specifically, FEC needs to handle subtle nuance between labels, which can be complex and confusing. Most existing models only address text classification problem in the euclidean space, which we believe may not be the optimal solution as labels of close semantic (e.g., \textit{afraid} and \textit{terrified}) may not be differentiated in such space, which harms the performance. In this paper, we propose \modelname, a novel framework that can integrate hyperbolic embeddings to improve the FEC task. First, we learn label embeddings in the hyperbolic space to better capture their hierarchical structure, and then our model projects contextualized representations to the hyperbolic space to compute the distance between samples and labels. Experimental results show that incorporating such distance to weight cross entropy loss substantially improves the performance with significantly higher efficiency. We evaluate our proposed model on two benchmark datasets and found 4.8\% relative improvement compared to the previous state of the art with 43.2\% fewer parameters and 76.9\% less training time. Code is available at \url{https://github.com/dinobby/HypEmo}.

\end{abstract}

\section{Introduction}
Fine-grained classification is a challenging yet important task that involves
differentiating subtle distinctions in a label set. For instance, in image
classification, classifying cars, planes, and other vehicles is
\textit{coarse-grained} classification, whereas distinguishing
models from cars is \textit{fine-grained} classification.
In NLP, sentiment analysis, which attempts to classify positive/negative
sentiments, is an example of coarse-grained text classification. Human emotions,
however, exhibit more complexity. For example, the six type of basic
emotion~\cite{ekman_basic_emotion} include \textit{happiness, sadness, fear,
disgust, anger}, and \textit{surprise}, and show finer distinctions of positive and
negative classes. Moreover, complex interactions exists in human emotion as
different type of emotions can have subtle differences, for instance
\textit{ashamed} and \textit{guilty}. This makes fine-grained emotion classification (FEC) 
challenging not only
because of the increased number of classes, but also because of the increased similarity between classes.
For instance, the current finest emotion classification datasets
contain up to 27 and 32 classes of
emotion~\cite{rashkin-etal-2019-towards,demszky-etal-2020-goemotions},
respectively. Classes such as \textit{furious} and \textit{angry} in these
fine-grained datasets are far more difficult to differentiate than
\textit{happy} and \textit{sad}. 
\begin{figure}[t!]
\includegraphics[width=\columnwidth]{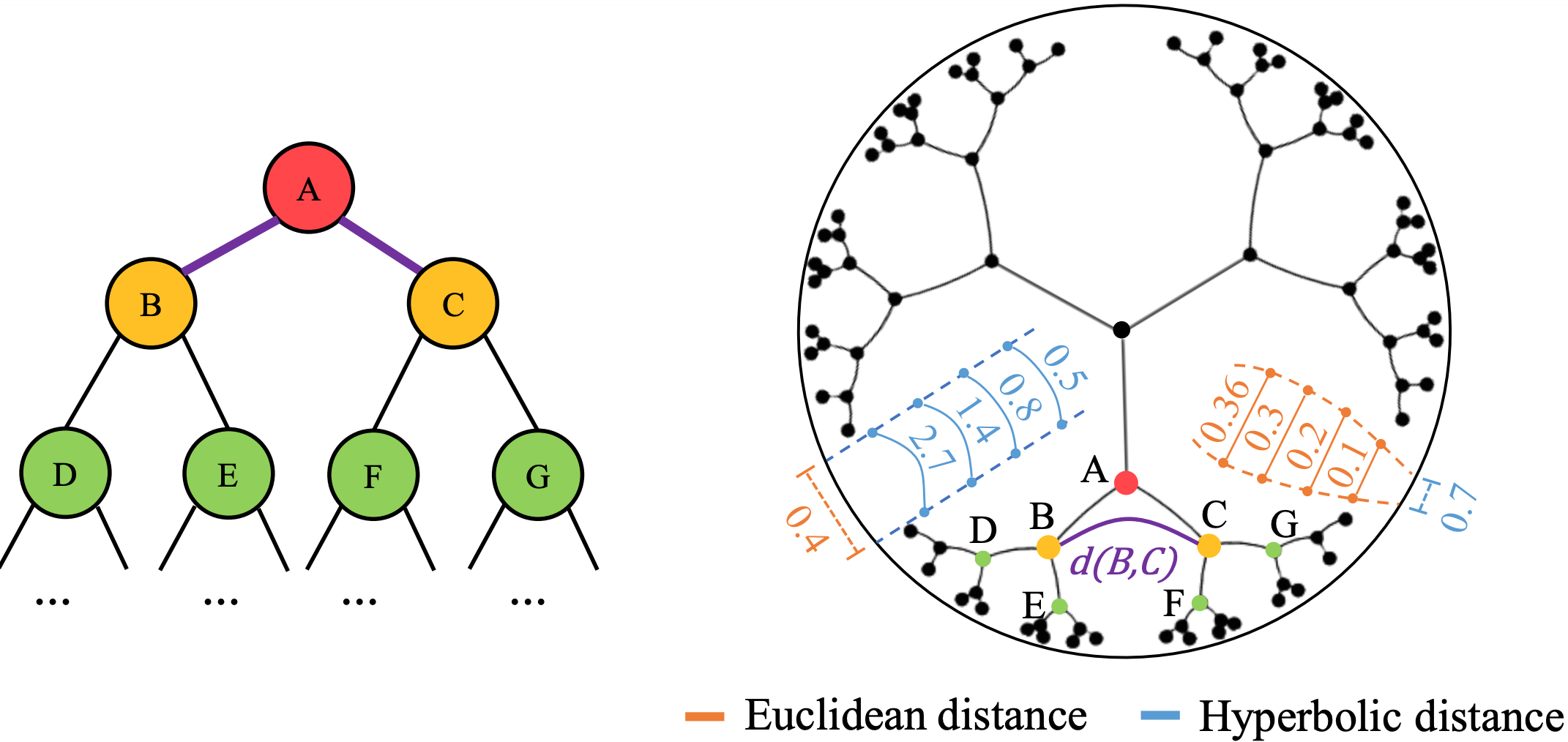}
\caption{\label{fig:intro_poincare}Properties of hyperbolic space.
The hyperbolic distance between two points grows dramatically as they come close to
the border, converse to the Euclidean distance. Moreover, the hyperbolic
distance is an analog of   
tree distance, which is $d(B,C) \approx d(B,A) + d(A,C)$. }
\end{figure}
However, detecting fine-grained emotion is useful in a variety of situations.
For example, in a dialog generation system, understanding the user's
fine-grained emotion could facilitate more empathy in the responses, which might
increase conversation engagement~\cite{roller-etal-2021-recipes}. 
Emotions also play an important role in people's lives, affecting how they make decisions and how they interact with machines. Therefore, the finer we can classify, the more information we can collect to model users' behaviors.

Existing text classification methods often use pre-trained language models such as
BERT~\cite{devlin-etal-2019-bert} to generate a contextualized representation,
and fine-tune them primarily in Euclidean space for downstream tasks. 
However, such a space is limited, as some confusing pairs have
nearly identical meanings (e.g., \textit{furious} and \textit{angry}), 
and forcing them to be separated in the latent space may harm the performance
by overfitting on training data.
The complex nature of emotions can be expressed in a hierarchical way, consisting of three levels~\cite{parrott2001emotions}.
For instance, \emph{Joy}, \emph{Contentment}, and \emph{Pleasure} are primary, secondary, and tertiary emotions, respectively.
Meanwhile, learning embeddings in hyperbolic space is becoming more popular due to its superior
ability to capture hierarchical information~\cite{NIPS2017_poincare_hierarchical, 
ganea2018hyperbolic,chami2019hyperbolic_gcn,liu2019hyperbolic_gnn}.
Figure~\ref{fig:intro_poincare} demonstrates a tree embedded to hyperbolic
space. Tree nodes closer to the root are embedded near the origin,
and nodes closer to leaves are placed closer to the boundary. The
main merit of this space is that as the distance from the origin rises, the
amount of space in hyperbolic space grows
exponentially~\cite{cho2019large,lopez-strube-2020-fully,2021_hyper_survey}.
Intuitively, tree-like structures also expand the number of nodes as the
distance increases from the root, which is consistent with the mathematical basis
of hyperbolic geometry. This is also reflected in 
the hyperbolic distance (Eq.~\ref{eq:poincare_distance}), 
which resembles
the distance between two nodes in a tree.

In this work, we propose \modelname, which integrates label embedding trained
in hyperbolic space with a RoBERTa model fine-tuned in Euclidean space.
Specifically, we first learn hierarchical-aware label embeddings in 
hyperbolic space, and then project the representation output by RoBERTa onto
the same space to derive the distance between a text representation and its
corresponding label. This distance is then used to weight standard cross
entropy loss, making the projection of text representation as close to its
label as possible, in hyperbolic space. 
Results on two challenging datasets,
GoEmotions~\cite{demszky-etal-2020-goemotions} and
EmpatheticDialogs~\cite{rashkin-etal-2019-towards}, demonstrate the superiority of the
proposed model. Also, we find that \wsmodelname performs best when the
label structure is more complex, or the inter-class relationship is more
ambiguous.
To sum up, the contributions of this paper are threefold:
\begin{itemize}
	 \item We leverage the merits of hyperbolic geometry to learn better
	 representations in both hyperbolic and Euclidean space. 
	 \item We propose the novel \wsmodelname framework along with a simple yet
	 effective objective function to address the FEC task.
	 \item Empirically, the proposed model outperforms existing methods, and
	 is even comparable with systems that utilize external knowledge or data
	 augmentation techniques. 
\end{itemize}

\section{Related Work}
\subsection{Fine-grained Classification}
Most of the literature addresses fine-grained text classification in Euclidean space.
\citet{khanpour-caragea-2018-fine} propose combining lexicon-based features to
detect fine-grained emotions in online health posts.
\citet{yin-etal-2020-sentibert} demonstrate that pre-trained models can
learn compositional sentiment semantics with self-attention applied to a
binary constituency parse tree and transfer to downstream sentiment
analysis tasks.
\citet{coarse2fine} propose utilizing generative language models for
fine-grained classification on coarsely annotated data.
\citet{suresh-ong-2021-negatives} propose label-aware contrastive loss (LCL),
which estimates the model confidence for each sample, and use this to weight
supervised contrastive loss~\cite{NEURIPS2020_supcon}. 
All of the above-mentioned work addresses FEC task primarily on the euclidean space, while we argue that some emotion with close semantics are not separable in this latent space. In our work, we integrate hyperbolic space to address this issue that improves FEC task.


\subsection{Hyperbolic Geometry}
A hyperbolic space is a non-Euclidean space for which the parallel
postulate does not hold~\cite{2021_hyper_survey, dhingra-etal-2018-embedding-text-hyper}.
The parallel postulate asserts that for every line~$L$ and 
point~$P$ not on $L$, there is a unique line that passes through $P$ that shares the same
plane with $L$ and $P$ and yet does not intersect with $L$. Without this
postulate, familiar mathematical properties in Euclidean space are
different. For example, in hyperbolic space, there can be more than one
line parallel to line~$L$ that goes through a point~$P$ not on $L$. Also,
whereas the distance between two points is a straight line in Euclidean space, this can
be generalized as a geodesic $\in [0, 1]$ which is the minimized distance
between two points. Moreover, the hyperbolic distance grows exponentially as
the points approach the boundary, making it more spacious than
Euclidean space given the same dimensions. These properties suit the nature of a tree-like structure, as the number of nodes grows exponentially
when the depth increases. In our work, we leverage the nature of hyperbolic
geometry to better capture label hierarchy, and propose improving the FEC task by jointly learning representations in both
Euclidean and hyperbolic space.

\subsection{Poincar\'{e} Embeddings}
The Poincar\'{e} ball model~\cite{Cannon_hyperbolicgeometry} is commonly
adopted in hyperbolic neural networks (HNN) and representation learning
research due to its differentiable distance
function~\cite{dhingra-etal-2018-embedding-text-hyper,NIPS2017_poincare_hierarchical}.
The Poincar\'{e} ball model is a Riemannian manifold that can be formulated as
$(B, g_x^b)$ with a Riemannian metric $g_x^b = \lambda_x^2 g^E$, where
$\lambda_x^2 = \frac{2}{1 - \Vert x \Vert^2}$ is called the \textit{conformal
factor}, and $g^E = I_n$ is the Euclidean metric tensor. The Riemannian metric
defines the geometric properties of a space, such as distances, angles, or curve
length. For example, Euclidean space is a manifold with zero curvature, and
the distance between two points can be written as $d(x, y) =
\sqrt{\sum_i{(x_i-y_i)}^2}$. The Poincar\'{e} ball model, on the other hand, is
a manifold with a constant negative curvature, where $B = \{x \in \mathbb{R}^n :
\Vert x \Vert < 1\}$ is an unit ball.
In natural language processing, researchers have applied such embeddings to
tasks as varied as fine-grained entity
typing~\cite{lopez-strube-2020-fully}, text classification~\cite{cho2019large}, and language modeling~\cite{dhingra-etal-2018-embedding-text-hyper}. 

\section{Methodology}
\begin{figure*}
\includegraphics[width=0.9\textwidth]{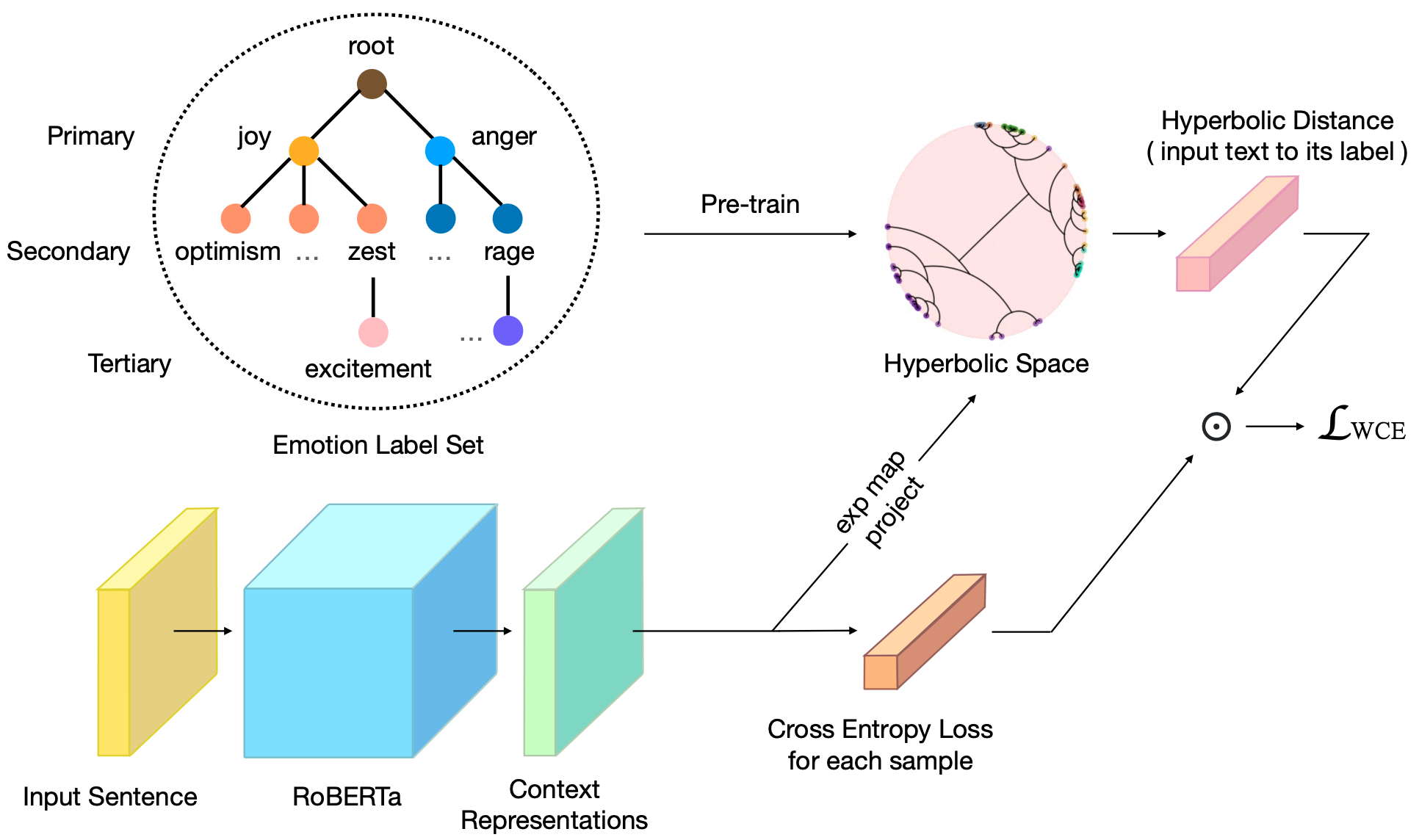}
\centering
\caption{\label{fig:model}The overall architecture of proposed framework. Label embeddings are pre-trained on the hyperbolic space to capture the hierarchical information, and the input sequence is encoded by RoBERTa to generate contextualized representations $h_i$ and cross entropy loss for each sample. $h_i$ will be simultaneously projected to the hyperbolic space to derive the hyperbolic distance $w_i$ to its label, and $w_i$ will be used to weight the standard cross entropy loss.}
\centering
\end{figure*}

In this section, we describe the proposed \wsmodelname in detail. Fig.~\ref{fig:model} illustrates its workflow. 

\subsection{Euclidean Sequence Representations}
Given the input sequence $x_i = \{x_i^1, x_i^2, \dots , x_i^k\}$ with $k$ tokens, text encoder generates the representation of the input sequence $h_i$. The text encoder itself is model agnostic, which could be any transformer-like model or even trained directly on the hyperbolic space. Here we use $\mathrm{RoBERTa}_\mathrm{base}$ in particular, since its excellence of generating contextualized representations. We also discuss the generalization of BERT-style models as the text encoder in the experiment. As convention, we take the hidden states corresponding to $\mathrm{[CLS]}$ token as the sequence representation $h_i$. 

\subsection{Hyperbolic Projection}
We adopt the Poincar\'{e} ball model of hyperbolic space. Let $B$ be the Poincar\'{e} ball model, and the associated tangent space denoted as $\tau_x B$, we use exponential map $ exp_x: \tau_x B \rightarrow B $, $\forall x \in B$, to project points from euclidean space to the hyperbolic space:
\begin{equation}
\label{eq:exp_map}
    \textrm{exp}_x(v) = x \oplus \textrm{tanh}(\frac{{\lambda_x \Vert v \Vert}}{2})\frac{v}{\Vert v \Vert}
\end{equation}
On the contrary, we could use logarithmic map to project points back to the euclidean space if needed:

\begin{equation}
    \textrm{log}_x(y) = \frac{2}{\lambda_x}\textrm{tanh}^{-1}(\Vert -x \oplus y \Vert)\frac{-x \oplus y}{\Vert -x \oplus y \Vert}
\end{equation}
where $v \neq 0$ and $y \neq x$ is the tangent vector, $\lambda_x = \frac{2}{1-\Vert x \Vert^2}$ is the conformal factor, and $\oplus$ is the Möbius addition:
\begin{equation}
    x \oplus y = \frac{(1+2\langle x,y \rangle + \Vert y \Vert^2)x + (1 - \Vert x \Vert^2)y}{1+2 \langle x,y \rangle + \Vert x \Vert^2 \Vert y \Vert^2}
\end{equation}
With exponential and logarithmic map, we can project embeddings from euclidean space to hyperbolic space or vice versa, and hence allowing us to take advantage on both spaces. Specifically, most of the well known language models are pre-trained on euclidean space which are powerful and easy to use, while in the hyperbolic space we can better model the label inventories in order to boost the performance.

\subsection{Hyperbolic Label Embeddings}
\label{sec:hyper_label_embedding}
To fully utilize the hierarchy of label set, we train label representations on the hyperbolic space. In this stage, our goal is to learn representations for each class $ \mathcal{E} = \{ e_1, e_2, \dots, e_m \}$, where $e_i$ is the hyperbolic label embeddings $ \in \mathbb{R}^{h_d}$, $h_d$ is the dimension of hyperbolic space, and $m$ is the number of classes. The label set can be represented as a set of tuples, indicating the parent-children relationship between nodes: $\mathcal{D} = \{(u,v)\}$ where $u$ is the parent of $v$. For datasets which does not contain parent-children relationship, we follow the parrot's emotion model~\cite{parrott2001emotions} to form $\mathcal{D}$, which has at most three levels of hierarchy. For the objective, we follow previous work~\cite{NIPS2017_poincare_hierarchical, ganea2018hyperbolic} that maximizes the distance between unrelated samples using negative sampling:
\begin{equation}
    \mathcal{L}_{label} = -\sum_{(u,v) \in \mathcal{D}} log \frac{e^{-d(u,v)}}{\sum_{v^{'} \in \mathcal{N}(u) \cup \{v\}}e^{-d(u,v^{'})}}
\end{equation}
where $\mathcal{N}(u) = \{v: (u,v) \not\in \mathcal{D}, v \neq u \}$ is the negative sample set and $d(u,v)$ is the distance between two points calculated by Eq.~\ref{eq:poincare_distance}. We use Riemammian Adam~\cite{becigneul2018riemannian_adam} for the optimization. After that, hyperbolic embeddings for each label is ready for use in the next step.

\subsection{Label-Aware Hyperbolic Weighted Loss}
\label{sec:hyperbolic_weighted_loss}
Cross entropy loss is commonly used in classification task. It assumes that every instance's negative log probability contributes equally. Usually, classifying a sample to be \textit{furious} when the ground truth is \textit{angry}, is more forgivable than classifying a sample to be \textit{joy}. However, it is the subtle difference between those confusable pairs such as \textit{angry} and \textit{furious} that makes fine-grained classification task challenging. 
In our work, we incorporate hyperbolic distance to enhance learning efficacy. To be more specific, we expect the confusable pairs that shares almost identical semantics can be well seperated on the hyperbolic space, and jointly update the model lies in the euclidean space. Formally, the pre-trained hyperbolic label embedding set (See Sec.\ref{sec:hyper_label_embedding}) is denoted as $ \mathcal{E} = \{ e_1, e_2, \dots, e_m \}$. Each instance $\{x_i, y_i\}$ contains a pair of sequences and labels where $y_i \in \mathcal{M}$ and $\mathcal{M}$ denotes the label set with $|\mathcal{M}| = m$. Given the sequence $x_i$, $h_i$ is extracted from the text encoder, and the logit $c_i \in \mathbb{R}^m$ is obtained by further passing $h_i$ through a linear layer: $c_i = \textrm{MLP}(h_i)$.
The standard cross-entropy loss is expressed mathematically as follows. 
\begin{equation}
\mathcal{L}_\mathcal{CE} = \frac{1}{N} \sum_{i = 1}^N-log\frac{exp(c_i^{y_i})}{\sum_{j = 1}^K exp(c_i^j)}
\end{equation}

The length of the geodesic, i.e., the distance between two points in a Poincar\'{e} ball is given by:
\begin{equation}
\label{eq:poincare_distance}
d(x_i, y_i) = \textrm{cosh}^{-1}(1 + 2 \frac{\Vert x_i -y_i \Vert^2}{(1 - \Vert x_i \Vert^2)(1 - \Vert y_i \Vert^2)})
\end{equation}

Now we can project the text representation generated from the encoder to the hyperbolic space using Eq.~\ref{eq:exp_map}, and calculate the distance $w \in \mathbb{R}$ of text representation and the label embeddings, which are both on the hyperbolic space.
We expect the embeddings for both input sequence and its label to be as close as possible, which means the distance $w$ is expected to be minimized. A simple way is to integrate $w$ into $\mathcal{L}_\mathcal{CE}$ by multiplying: 
\begin{equation}
    \mathcal{L}_\mathcal{WCE} =  \sum_{i = 1}^N-w_i log \frac{exp(c_i^{y_i})}{\sum_{j = 1}^K exp(c_i^j)}
\end{equation}

In this way, $w$ can be viewed as a weight to penalize pairs that have larger distance in the hyperbolic space. Our goal here is to jointly learn embeddings in both Euclidean and hyperbolic space to boost the performance. By taking $w$ as a weight to sum up cross entropy loss, the main merit is that it is easy to implement without carrying all optimization operation on the hyperbolic space, while allowing the whole framework to be updated jointly. We will also discuss the scenario when fully using hyperbolic neural networks in Sec.~\ref{sec:main_results}.

\section{Experiment}
\subsection{Datasets}
We evaluate our method on two datasets:
GoEmotions~\cite{demszky-etal-2020-goemotions}, and Empathetic Dialogues~\cite{rashkin-etal-2019-towards}. 
Given our primary objective of fine-grained emotion classification, which requires distinguishing labels with subtle differences in meaning, we choose GoEmotions and Empathetic Dialogues for evaluation. These datasets are considered the most challenging datasets, as they contain a larger number of emotion labels with similar semantics.
Below we give the descriptions of these datasets.
GoEmotions is composed of comments from
Reddit~\cite{demszky-etal-2020-goemotions}. The total number of samples is 54k,
and each sample is annotated with one or multiple labels among 27 emotions and
neutral. To ensure a fair comparison with previous work~\cite{suresh-ong-2021-negatives}, 
we use only single-labeled samples and
exclude neutral ones. 
The training/validation/test split of the remaining dataset is 23,485\,/\,2,956\,/\,2,984.

Empathetic Dialogues~\cite{rashkin-etal-2019-towards} consists of conversations
with single emotion labels.  The situation was written by a speaker given an
emotion label. The listener was to respond with an empathetic utterance.
The process could go on for up to six turns.  Since the situation was written
based on the provided emotion, we used the situation as the model's input, 
following~\citet{suresh-ong-2021-negatives}. The dataset contains 24,850
conversations labeled among 32 emotions. The training/validation/test split of
the dataset is 19,533\,/\,2,770\,/\,2,547, respectively. Below, we use GE to
represent GoEmotions and ED to represent Empathetic Dialogues.






\subsection{Experiment Settings and Baselines}
We compare the proposed \wsmodelname primarily with three categories of strong
baselines:

\noindent\textbf{General pre-trained language models.} 
We compared against BERT~\cite{devlin-etal-2019-bert}, RoBERTa~\cite{roberta},
and ELECTRA~\cite{clark2020electra}. These models are pre-trained on large unlabeled corpora. They generate high-quality representations, and all perform well on text classification tasks. We also compare different size of these pre-trained models, denoted as $\mathrm{base}$ and $\mathrm{large}$.

\noindent\textbf{Label Embedding-aware models.}
\citet{suresh-ong-2021-negatives} propose label-aware contrastive loss (LCL),
which weights each negative sample differently. Specifically, more confusable pairs contribute more to the objective function; this yields promising results on fine-grained text classification.
In addition, we compare against HiAGM~\cite{jie2020hierarchy}, the strongest hierarchy-aware text classification model with source code publicly available.
Lastly, we implement a baseline called LabelEmb, which encodes the label description (i.e., the definition of emotions) to derive label embeddings, and train the model on the Euclidean space with the rest setting same as \modelname.

\noindent\textbf{Hyperbolic classification models.}
We also compared with models trained in hyperbolic space for classification, including (1)~Hyperbolic SVM (HSVM) proposed by~\citet{cho2019large}, which
generalizes the support vector machine to hyperbolic space, and (2)~Hyperbolic
Neural Model (HNN) proposed by~\citet{ganea2018hyperbolic}, a hyperbolic GRU that performs all necessary operations in hyperbolic space to train a neural
network. (3)~Hyperbolic Interaction Model (HyperIM)~\cite{chen2020hyperbolic} jointly learns word and label embeddings, and measure the similarities in the Poincor\'{e} disk to aggregate input representations.
(4)~\textsc{HiddeN}~\cite{chatterjee-etal-2021-joint} is a framework which does not assume the label hierarchy is known. It also proposed to learn the label embedding jointly in an end-to-end fashion.
For a fair comparison, we set the word dimension of the hyperbolic 
space to 100, the same as 
the dimension we use.  

\noindent\textbf{Evaluation metrics.}
Following~\citet{suresh-ong-2021-negatives}, we use accuracy and weighted F1 as
the evaluation metrics. Weighted F1 takes into account the number of samples in each
class, and weights the macro F1 by this ratio. This can be expressed as
\begin{equation}
    F1_{\mathit{weighted}} = 2 \sum_c \frac{n_c}{N} \frac{P_c \times R_c}{P_c + R_c},
\end{equation}
where $n_c$ is the number of samples in class~$c$, $N$ is the number of total
samples, and $P_c$ and $R_c$ are the precision and recall for class~$c$, respectively.

\noindent\textbf{Implementation Details.} \wsmodelname is a encoder-agnostic framework which can easily adapt to different kinds of text encoders. 
In the experiment, we use pre-trained $\mathrm{RoBERTa}_\mathrm{base}$ as the backbone, which has 12 layers with a hidden size of 768. 
During training, we applied the Adam optimizer in Euclidean space
with a learning rate of $10^{-5}$ and a weight decay of $0.01$. 
By contrast, we utilized Riemannian Adam~\cite{becigneul2018riemannian_adam} 
to train our label
embeddings in hyperbolic space with a learning rate of $0.01$. The dimension of hyperbolic label embedding is set to 100, which is searched from \{2, 10, 50, 100, 250\}. Other implementation details can be found in our code.

\subsection{Main Results}
\label{sec:main_results}
\noindent\textbf{Baseline comparison.} To demonstrate the effectiveness of \modelname, we conduct experiments to compare the performance of different models. The comparison is shown in Table~\ref{tab:main_results}. Firstly, we compare \wsmodelname with general pre-trained language models. Among them, $\mathrm{RoBERTa}_\mathrm{large}$ performs the best, while \wsmodelname outperforms it on weighted F1 by 2.8\% on ED and 1.1\% on GE. It is also worth mentioning that \wsmodelname has considerably smaller parameter size compared with  $\mathrm{RoBERTa}_\mathrm{large}$ (125M v.s. 355M), resulting in significantly lower training and inference time.
This indicates the effectiveness of the proposed label-aware hyperbolic embeddings and the strategy to weight the standard cross entropy loss by hyperbolic distance.  

Secondly, we compare with label-aware system. Since LCL is the previous state-of-the-art, we mainly compare the efficiency with it ($\Delta$ in the left of Table~\ref{tab:main_results}). Results show that our proposed method outperforms LCL with much higher efficiency. This is because LCL augments data by using the synonym replacement technique, which doubles the size of data. Also, they use two encoders to train the main classifier and a weighting network, which doubles the parameter size. In contrast, \wsmodelname uses single encoder and uses only the original samples without any data augmentation method, and still out-wins LCL by 2.8\% and 2.5\% absolute F1 score on ED and GE, respectively. 
Moreover, Although HiAGM take into account the label hierarchy, it utilize RNN architecture, making it less efficient and underperforming \wsmodelname by a large margin.
Lastly, \wsmodelname performs better than LabelEmb, which LabelEmb calculates the weighted loss in the Euclidean space. This again demonstrates the efficacy of our proposed hyperbolic space integration. 

Also, we notice that \wsmodelname works better than models that are fully trained on hyperbolic space, which indicates the benefits of jointly learning the hyperbolic label embedding and fine-tuning $\mathrm{RoBERTa}_\mathrm{base}$ in a hybrid space settings. This hybrid setting could benefit from both the power of pre-trained language model and the strength of hyperbolic space to capture hierarchical information. To sum up, we could achieve better results compared to previous works without increasing data size or model parameters, which is more effective and efficient.  

\begin{table*}[t]
    \small
    \centering
    \begin{tabular}{c c c c c | c c c c }
    \bottomrule
    \\[-1em]
         \multicolumn{5}{c}{} & \multicolumn{2}{c}{ Empathetic Dialogues} & \multicolumn{2}{c}{ GoEmotions} \\ \\[-1em] \hline \\[-1em]
         \multicolumn{1}{c}{} & \textbf{\#Params (\textdownarrow)} & $\xi_{tr}$ (\textdownarrow) & $\xi_{conv}$ (\textdownarrow) & $\xi_{inf}$ (\textdownarrow)& \textbf{ Acc} (\textuparrow)& \multicolumn{1}{c}{\textbf{ F1} (\textuparrow)}  & \textbf{ Acc} (\textuparrow) & \multicolumn{1}{c}{\textbf{F1} (\textuparrow)} \\ \\[-1em] \toprule 
      
         
         \multicolumn{1}{c}{$\mathrm{BERT}_\mathrm{base}$} & 110M & 96.7 & 290.0 & 1.1 & 50.4 \footnotesize{$\pm$0.3} & \multicolumn{1}{c}{51.8 \footnotesize{$\pm$0.1}} & 60.9 \footnotesize{$\pm$0.4} & \multicolumn{1}{c}{62.9 \footnotesize{$\pm$0.5}} \\

         \multicolumn{1}{c}{$\mathrm{RoBERTa}_\mathrm{base}$} & 125M & 99.3 & 297.9 & 1.2 & 54.5 \footnotesize{$\pm$0.7}  &  \multicolumn{1}{c}{56.0 \footnotesize{$\pm$0.4}}  &  62.6 \footnotesize{$\pm$0.6} &  \multicolumn{1}{c}{64.0 \footnotesize{$\pm$0.2}}\\
         
         \multicolumn{1}{c}{$\mathrm{ELECTRA}_\mathrm{base}$} & 110M & 97.6 & 292.7 & 1.1 & 47.7 \footnotesize{$\pm$1.2} & \multicolumn{1}{c}{49.6 \footnotesize{$\pm$1.0}} & 59.5 \footnotesize{$\pm$0.4} & \multicolumn{1}{c}{61.6 \footnotesize{$\pm$0.6}} \\
         
         \multicolumn{1}{c}{$\mathrm{BERT}_\mathrm{large}$} & 340M & 181.0 & 362.0 & 3.5 & 53.8 \footnotesize{$\pm$0.1} & \multicolumn{1}{c}{54.3 \footnotesize{$\pm$0.1}} & 64.5 \footnotesize{$\pm$0.3} & \multicolumn{1}{c}{\underline{65.2} \footnotesize{$\pm$0.4}} \\

         \multicolumn{1}{c}{$\mathrm{RoBERTa}_\mathrm{large}$} & 355M & 185.7 & 371.4 & 3.7 & 57.4 \footnotesize{$\pm$0.5}  &  \multicolumn{1}{c}{\underline{58.2} \footnotesize{$\pm$0.3}}  &  \underline{64.6} \footnotesize{$\pm$0.3} &  \multicolumn{1}{c}{\underline{65.2} \footnotesize{$\pm$0.2}}\\
         
         \multicolumn{1}{c}{$\mathrm{ELECTRA}_\mathrm{large}$} & 335M & 179.9 & 539.8 & 3.5 & 56.7 \footnotesize{$\pm$0.6} & \multicolumn{1}{c}{57.6 \footnotesize{$\pm$0.6}} & 63.5 \footnotesize{$\pm$0.3} & \multicolumn{1}{c}{64.1 \footnotesize{$\pm$0.4}} \\
         \hline
                  
         
         \\[-8pt]
         \multicolumn{1}{c}{$LCL^\dag$} & 220M & 421.7 & 1264.9 & 3.9 & \underline{59.1} \footnotesize{$\pm$0.4} & \multicolumn{1}{c}{\underline{58.2} \footnotesize{$\pm$0.5}} & \underline{64.6} \footnotesize{$\pm$0.2} & \multicolumn{1}{c}{63.8 \footnotesize{$\pm$0.3}}\\
         
         \multicolumn{1}{c}{HiAGM} & 15M & 2673.2 & 10692.9 & 8.6 & 47.8 \footnotesize{$\pm$0.6} & \multicolumn{1}{c}{50.2 \footnotesize{$\pm$0.7}} & 59.7 \footnotesize{$\pm$0.6} & \multicolumn{1}{c}{61.8 \footnotesize{$\pm$0.5}}\\
         
         \multicolumn{1}{c}{LabelEmb} & 125M & 103.6 & 518.1 & 1.1 & 55.1 \footnotesize{$\pm$0.7} & \multicolumn{1}{c}{56.2 \footnotesize{$\pm$0.5}} & 62.7 \footnotesize{$\pm$0.6} & \multicolumn{1}{c}{62.8 \footnotesize{$\pm$0.4}}\\
         \hline
         \\[-8pt]
         \multicolumn{1}{c}{HSVM} & 428K & 14.7 & 42.1 & 0.7 & 27.4 \footnotesize{$\pm$0.0} & \multicolumn{1}{c}{26.7 \footnotesize{$\pm$0.0}} & 23.6 \footnotesize{$\pm$0.0} & \multicolumn{1}{c}{22.3 \footnotesize{$\pm$0.0}}\\
         
         \multicolumn{1}{c}{HNN} & 5M & 15421.5 & 92529.1 & 17.5 & 41.2 \footnotesize{$\pm$0.9} & \multicolumn{1}{c}{42.0 \footnotesize{$\pm$0.8}} & 46.6 \footnotesize{$\pm$0.6} & \multicolumn{1}{c}{47.2 \footnotesize{$\pm$0.5}}\\

          \multicolumn{1}{c}{HyperIM} & 5M & 2266.3 & 6798.9 & 8.4 & 44.1 \footnotesize{$\pm$1.2} & \multicolumn{1}{c}{43.6 \footnotesize{$\pm$1.0}} & 50.2 \footnotesize{$\pm$0.9} & \multicolumn{1}{c}{49.7 \footnotesize{$\pm$0.7}}\\
          
          \multicolumn{1}{c}{\textsc{HiddeN}} & 11M & 13473.0 & 67364.8 & 16.6 & 42.9 \footnotesize{$\pm$1.4} & \multicolumn{1}{c}{44.3 \footnotesize{$\pm$1.1}} & 47.2 \footnotesize{$\pm$1.1} & \multicolumn{1}{c}{49.3 \footnotesize{$\pm$0.9}}\\
          
         \hline
         \\[-8pt]
          \multicolumn{1}{c}{\wsmodelname} & 125M & 97.6 & 585.8 & 1.2 & \textbf{59.6} \footnotesize{$\pm$0.3} & \multicolumn{1}{c}{\textbf{61.0} \footnotesize{$\pm$0.3}}  & \textbf{65.4} \footnotesize{$\pm$0.2}& \multicolumn{1}{c}{\textbf{66.3} \footnotesize{$\pm$0.2}} \\
          
          \multicolumn{1}{c}{$\Delta$} & -43.2\% & -76.9\% & -53.3\% & -68.5\% & +0.8\% & +4.8\% & \multicolumn{1}{c}{+0.8\%} & \multicolumn{1}{c}{+3.9\%}  \\
         \toprule

    \end{tabular}
	 \caption{Left: efficiency comparison (lower is better), where $\xi_{tr}$ denotes average training time for one epoch, $\xi_{conv}$ denotes the average time to achieve the best results, and $\xi_{inf}$ denotes the average inference time for the whole testing set. We evaluate efficiency for all systems on GE with batch size equals 16. Right: performance comparison (higher is better.) We conducted the experiment five times with different seeds, and report the average score with the standard deviation. The best (second-best) results are set in \textbf{bold} (\underline{underlined}). ``\dag'' denotes reproduced results from official implementation, and $\Delta$ represents the efficiency and performance gain compared with LCL, the previous state-of-the-art, in relative percentage.}
    \label{tab:main_results}
\end{table*}

\begin{table}[ht!]
\centering
\small
\begin{tabular}{clcc}
\bottomrule\\[-1em]
\multicolumn{1}{c}{Dataset} & \multicolumn{1}{c}{Model}   & \multicolumn{1}{c}{F1} & \multicolumn{1}{c}{F1 (+HypEmo)} \\\\[-1em] \toprule
\multirow{3}{*}{ED}  & $\mathrm{BERT}_\mathrm{base}$    & 51.8                   & \textbf{57.7}                    \\
                     & $\mathrm{RoBERTa}_\mathrm{base}$ & 56.0                   & \textbf{61.0}                    \\
                     & $\mathrm{ELECTRA}_\mathrm{base}$ & 57.6                   & \textbf{58.9}                    \\ \hline \\[-1em]
\multirow{3}{*}{GE}  & $\mathrm{BERT}_\mathrm{base}$    & 62.9                   & \textbf{65.3}                    \\
                     & $\mathrm{RoBERTa}_\mathrm{base}$ & 64.0                   & \textbf{66.3}                    \\
                     & $\mathrm{ELECTRA}_\mathrm{base}$ & 64.1                   & \textbf{65.7}                    \\ \toprule
\end{tabular}
\caption{Performance in terms of weighted F1 score when \wsmodelname is added on different encoders.}
\label{tab:encoder_analysis}
\end{table}


\noindent\textbf{Performance on different encoder.}
We apply \wsmodelname on top of different encoders to examine whether \wsmodelname is a model-agnostic method that could bring improvement regardless of the encoder being used. 
Table~\ref{tab:encoder_analysis} shows the results in terms of weighted F1 score. We observe that no matter which encoder is adopted, adding \wsmodelname leads to further improvements. For instance, applying \wsmodelname on $\mathrm{BERT}_\mathrm{base}$ enhances the performance by 5.9\% absolute percentage on ED, and the same phenomenon can be observed on $\mathrm{RoBERTa}_\mathrm{base}$ and $\mathrm{ELECTRA}_\mathrm{base}$ across two datasets. This verifies that \wsmodelname is model-agnostic and could be easily built on top of any text encoder to boost performance.

\subsection{Case Study}
\begin{figure}[ht!]
    \centering
    \includegraphics[width=0.475\textwidth]{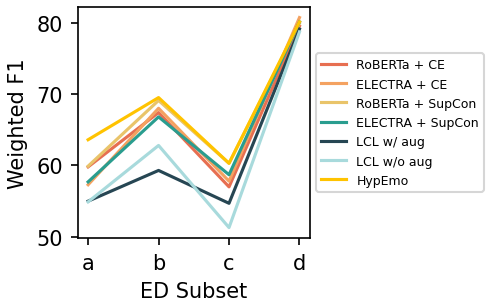}
    \caption{Case study for various ED subsets. We report F1 for brevity. CE is the standard cross entropy loss, and SupCon represents supervised contrastive loss~\cite{NEURIPS2020_supcon}.}
    \label{fig:ED_subset}
\end{figure}
Following~\citet{suresh-ong-2021-negatives}, we investigate the performance of
\wsmodelname when the label set contains pairs with close semantics. We compare
the proposed \wsmodelname with different objectives and encoders, and follow
the ED subsets selected by \citet{suresh-ong-2021-negatives}, which
include the most difficult sets chosen after enumerating all combinations that contain
four labels. These subsets are a: \{\emph{Anxious, Apprehensive, Afraid,
Terrified}\}, b: \{\emph{Devastated, Nostalgic, Sad, Sentimental}\}, c:
\{\emph{Angry, Ashamed, Furious, Guilty}\}, and d: \{\emph{Anticipating, Excited,
Hopeful, Guilty}\} from the ED dataset. We conducted the experiments with
RoBERTA and ELECTRA with standard cross entropy loss and
supervised contrastive loss~\cite{NEURIPS2020_supcon}. 
For supervised contrastive loss, we adopt back translation~\cite{sennrich-etal-2016-improving} to form the positive pairs,
and we view it as a strong competitor because we expect contrastive loss helps to learn better representations could hence improve the FEC task.
The result is shown in Fig~\ref{fig:ED_subset}. First, \wsmodelname outperforms all baselines by a large margin on the most difficult subset (a), which demonstrates the advantage of incorporating hyperbolic space when addressing fine-grained classification task. In particular, \wsmodelname beats the previous state-of-the-art, LCL, on this subset with a substantial improvement (54.9 v.s. 63.6.) Also, \wsmodelname surpasses models with cross entropy loss and supervised contrastive loss on a / b / c and comparably on d. As a / b / c are the top three most difficult sets, this result shows that label-aware hyperbolic weighted loss is conducive to separation under label sets which are more confusing. 
In addition, we compare with LCL with and without augmentation to ensure a fair comparison. The result shows
that \wsmodelname consistently outperforms LCL even with data augmentation that doubles the size of samples. 
In summary, the proposed model performs the best when the label set contains confusing classes. For the most challenging sets, \wsmodelname outperforms models trained with conventional cross entropy loss and supervised contrastive loss, and even the state-of-the-art, LCL, while being more sample-efficient. When the label set is simpler, \wsmodelname performs on par with the others.

\begin{figure*}[ht!]
\centering
\begin{subfigure}{.3\textwidth}
    \centering
    \includegraphics[width=4.75cm]{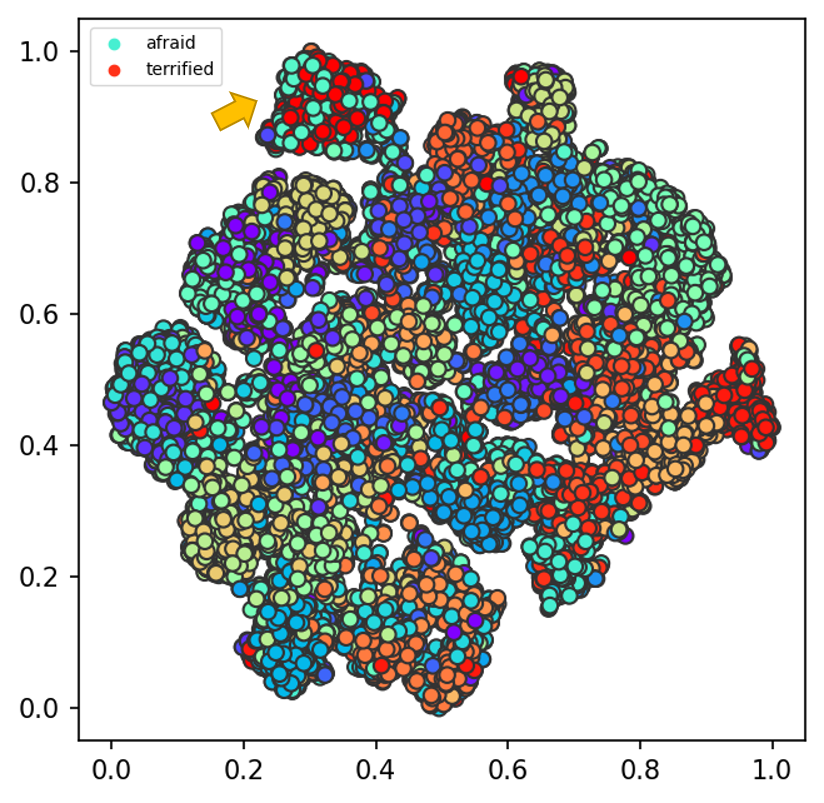}
    \caption{Standard cross entropy loss}
    \label{fig:tsne1}
\end{subfigure}
\begin{subfigure}{.3\textwidth}
    \centering
    \includegraphics[width=4.75cm]{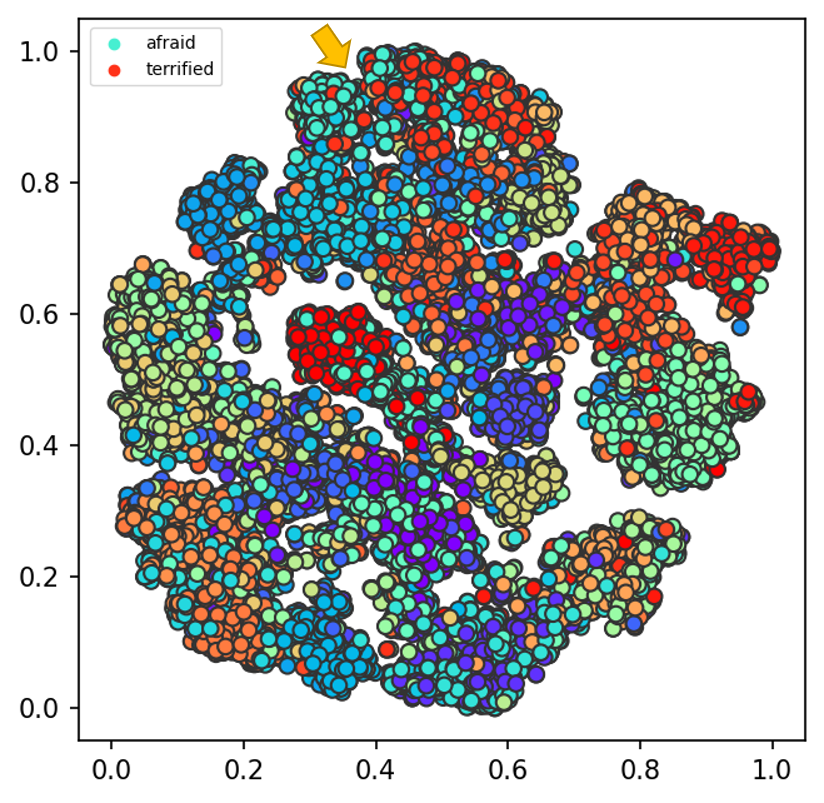}
    \caption{Supervised contrastive loss}
    \label{fig:tsne2}
\end{subfigure}
\begin{subfigure}{.3\textwidth}
    \centering
    \includegraphics[width=4.75cm]{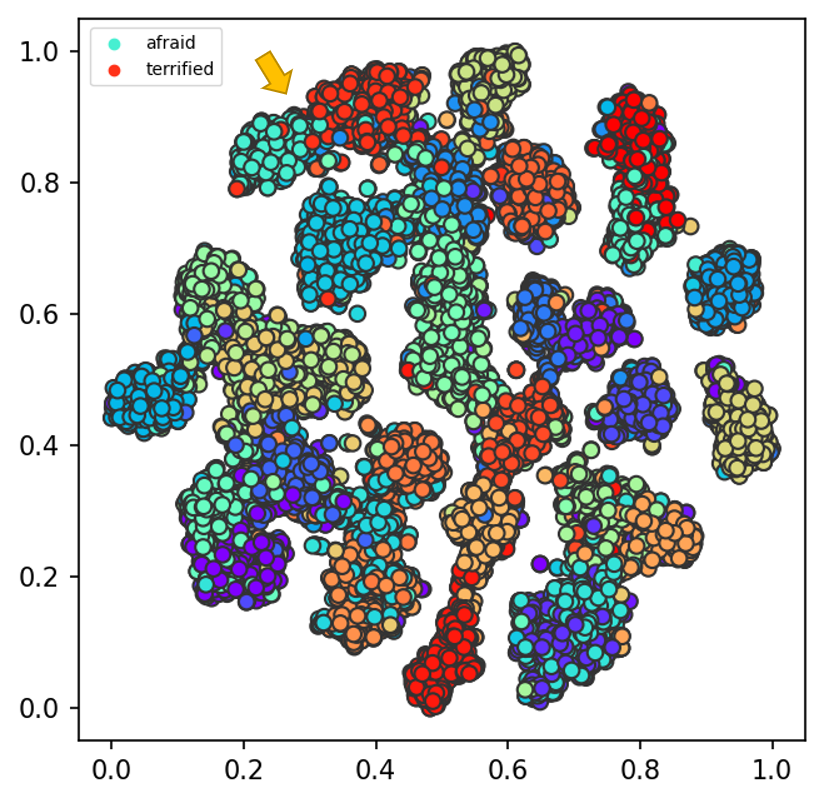}
    \caption{Hyperbolic weighted loss}
    \label{fig:tsne3}
\end{subfigure}
\caption{Representation generated from text encoder using different
objective functions. We take \emph{afraid} and \emph{terrified} as examples for semantically close labels, and we highlight their position with the yellow arrow. Evidently, our proposed label-aware hyperbolic weighted loss leads to larger
inter-class distances.}
\label{fig:visual}
\end{figure*}

\begin{figure}[]
\centering
\includegraphics[width=0.3\textwidth]{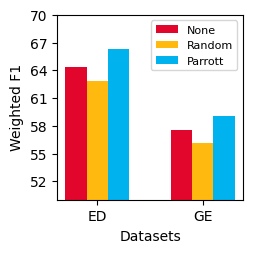}
\caption{Label hierarchy method vs.\ performance.}
\label{fig:hierarchy}
\end{figure}

\subsection{Effect of Label Hierarchy}
In this section, we investigate how the method used
to form the label hierarchy affects performance. We compare three settings: \emph{None}, \emph{Random} and \emph{Parrott}, as shown in Fig.~\ref{fig:hierarchy}. 
\textit{None} means that we abandon all hierarchy and distribute labels
uniformly on the Poincar\'{e} ball, and \textit{Random} means that
we randomly shuffle the correct order to make hierarchy meaningless. 
\textit{Parrott} is the emotion model proposed by~\citet{parrott2001emotions},
which defines a three-level emotion hierarchy. We observe that following 
an expert-defined hierarchy yields the best performance, better than without hierarchical
information, and the worst is shuffling to discard all hierarchical information.
Accordingly, we follow Parrott's emotion model to form the label hierarchy and learn the hyperbolic label embeddings.

\subsection{Visualization of Representations}
To better understand the effect of the proposed label-aware hyperbolic weighted
loss, we compare representations learned with different commonly used
objective functions. We trained on the ED dataset and projected
the embeddings generated from the text encoder onto 2-dimensional space with
t-SNE~\cite{van2008visualizing}. We compare standard cross entropy loss
(Fig.~\ref{fig:tsne1}), supervised contrastive loss~\cite{NEURIPS2020_supcon}
(Fig.~\ref{fig:tsne2}), and the proposed label-aware hyperbolic weighted loss
(Fig.~\ref{fig:tsne3}). For supervised contrastive loss, we also used
back translation as the augmentation to form
positive pairs. Then, standard cross entropy loss was summed with supervised
contrastive loss. As data augmentation doubles the data size, we
expect the supervision and the objective to lead to more a separated
representation.
In Fig.~\ref{fig:tsne1}, we observe indistinct clustering, as
many data points with different classes are mingled together, showing the
difficulties of fine-grained classification in which different classes
often share close semantics. In particular, \emph{afraid} and \emph{terrified} are confusing for models and their representations are mixed. 
With the use of supervised contrastive loss shown
in Fig.~\ref{fig:tsne2}, the clustering becomes somewhat clearer but at the
cost of increasing data.
Last, in Fig.~\ref{fig:tsne3}, the inter-class distance is
clearly larger than others, and the clusters are also more dispersed. 
Specifically, the representations of \emph{afraid} and \emph{terrified} are much more separated.
This shows the advantage of the proposed label-aware hyperbolic loss, which yields
better representations, even without the need for additional significant costs.
\section{Conclusion}
We propose \modelname, a novel framework that includes a
label-aware hyperbolic weighted loss to improve FEC task performance. By jointly
learning the representations in Euclidean and hyperbolic space,
we leverage hybrid settings that combine the power of large-scale
pre-trained language models and the mathematical characteristics of hyperbolic
space to capture the hierarchical property of classes and the nuanced differences
between them. With this design, the proposed method achieves 
state-of-the-art results in terms of weighted F1 on the GE and ED benchmark datasets.
We show that the proposed model works even better when the labels
are difficult to differentiate. 
Moreover, \wsmodelname outperforms methods that utilize data augmentation while being more efficient.

\section*{Limitations}
Although the proposed framework yields promising results on two fine-grained
emotion datasets---GoEmotions and Empathetic Dialogues---there remain
limitations, including:
(1)~To the best of our knowledge, there is no such fine-grained emotion
dataset in other languages. Although theoretically, our method should work fine on
languages other than English, we can only show the results in English.
(2)~The proposed method works best when the label structure contains
hierarchy, especially when the semantics of some labels are close and difficult to
distinguish. When the label structure is flat and independent, our method may
backoff to a conventional classification model.

\section*{Acknowledgement}
This work is supported by the National Science and Technology Council (NSTC) of Taiwan under grants 111-2221-E-001-021 and 111-2634-F-002-022.

\bibliography{custom}
\bibliographystyle{acl_natbib}

\appendix



\end{document}